\documentclass{article}

\usepackage{PRIMEarxiv}
\usepackage{natbib}
\usepackage[utf8]{inputenc} %
\usepackage{algorithm}
\usepackage{array}
\usepackage{algpseudocode}
\usepackage{subfig}
\usepackage{textcomp}
\usepackage{stfloats}
\usepackage{verbatim}
\usepackage[T1]{fontenc}    %
\usepackage{hyperref}       %
\usepackage{url}            %
\usepackage{booktabs}       %
\usepackage{amsfonts}       %
\usepackage{nicefrac}       %
\usepackage{microtype}      %
\usepackage{xcolor}         %

\usepackage{amsmath}
\usepackage{bm}
\usepackage{color}
\usepackage{graphicx}
\usepackage{makecell}
\usepackage{multirow}
\usepackage{tabularx}
\usepackage{mathtools}
\usepackage{pythonhighlight}
\usepackage{listings}

\definecolor{pink}{rgb}{0.858, 0.188, 0.478}
\usepackage{xspace}

\definecolor{commentcolor}{RGB}{110,154,155}   %

\title{More Women, Same Stereotypes: Unpacking the Gender Bias Paradox in Large Language Models
}

\author{
  Evan Chen, Run-Jun Zhan, Yan-Bai Lin, Hung-Hsuan Chen \\
  Computer Science and Information Engineering \\
  National Central University \\
  Taoyuan, Taiwan\\
  \texttt{evan07321tw@gmail.com, 111502543@cc.ncu.edu.tw, angelinalin2223@gmail.com, hhchen1105@acm.org} \\
}

\begin{document}
\maketitle

\begin{abstract}

Large Language Models (LLMs) have revolutionized natural language processing, yet concerns persist regarding their tendency to reflect or amplify social biases. This study introduces a novel evaluation framework to uncover gender biases in LLMs: using free-form storytelling to surface biases embedded within the models. A systematic analysis of ten prominent LLMs shows a consistent pattern of overrepresenting female characters across occupations, likely due to supervised fine-tuning (SFT) and reinforcement learning from human feedback (RLHF). Paradoxically, despite this overrepresentation, the occupational gender distributions produced by these LLMs align more closely with human stereotypes than with real-world labor data. This highlights the challenge and importance of implementing balanced mitigation measures to promote fairness and prevent the establishment of potentially new biases. We release the prompts and LLM-generated stories at GitHub.\footnote{\url{https://github.com/chia0403/Quantifying_Occupational_Stereotypes_in_Large_Language_Models_Through_Open-Ended_Prompts}}

\end{abstract}

\keywords{LLM \and Gender Bias \and AI fairness \and Human-AI alignment}

\section{Introduction}

Large Language Models (LLMs) have rapidly become integral to natural language processing and decision support. Despite their capabilities, these models may inadvertently preserve or even amplify social biases rooted in training data. Accurately quantifying these biases is essential to ensure their fair and responsible deployment.

Previous studies have used various methods to assess LLMs' gender biases in professional contexts. Some approaches involve structured decision-making tasks, such as choosing candidates from lists of names tied to different demographic groups~\citep{nghiem-etal-2024-gotta, an-etal-2024-large}. Other strategies test linguistic biases using prompts with ambiguous pronouns, e.g., ``The doctor phoned the nurse because she was late''~\citep{10.1145/3582269.3615599}. Although effective, these methods are limited to predefined scenarios, risk oversimplifying biases, and may reveal the evaluation's intent, potentially allowing LLMs to adapt strategically.

To address these shortcomings, we propose a simple evaluation strategy that avoids predefined scenarios and conceals the intent of the test. Specifically, we request LLMs to generate a story opening and setting for a character in a specific occupation and analyze the gender cues in the output. We execute the exact prompt multiple times for each occupation and evaluate the gender distribution across repeated outputs. Thus, we eliminate the need for strictly designed scenarios or carefully tailored prompts, allowing biases to appear naturally within free-form storytelling and avoiding directing the model's attention toward gender-related considerations, providing an authentic view of occupational gender stereotypes.

Based on this new evaluation strategy, we observe the following. First, LLMs exhibit an overrepresentation of female characters in their generated stories, with 35 of 106 occupations showing female dominance ($80\%$ or more characters are female) across the 10 LLMs tested, whereas male characters did so in only 5 out of 106. Interestingly, this female dominance is not limited to traditionally stereotypical female-associated occupations (Section~\ref{subsec:female_overrepresentation}). A possible reason for this pattern is that alignment techniques, such as supervised fine-tuning (SFT) and reinforcement learning from human feedback (RLHF) employed by model developers, may have influenced the models towards increased female representation. This effort could lead to the overrepresentation of female characters in a wide range of occupations, not only those traditionally associated with women. We validated this assumption by comparing the models with and without SFT and RLHF (Section~\ref{subsec:GPT2-xl}). Second, the rankings of occupations by gender ratios in LLM outputs align more closely with the perceived gender associations of occupations (as measured by human ratings) than the actual gender distributions in the labor force. This indicates that LLMs are more likely to mirror societal perceptions of gender roles than real-world demographic distributions, potentially amplifying biases rooted in societal stereotypes (Section~\ref{subsec:rankings_vs_human_judgment}). Although this observation may seem contradictory to the first conclusion, it is not. The increased representation of females across all categories appears to lead many occupations to become female-dominated. However, when occupations are ranked by gender composition ratios, the resulting sequence still strongly aligns with human perceptions of gender associations.

\section{Related Work} \label{sec:rel-work}

\subsection{Evaluating Bias by LLMs' Decisions on Strictly Designed Scenarios}

Several studies have asked LLMs to make decisions or recommendations, allowing researchers to assess biases in these results~\citep{nghiem-etal-2024-gotta, an-etal-2024-large, Salinas_2023, haim2024whatsnameauditinglarge}. For example, \citet{nghiem-etal-2024-gotta} prompted LLMs to select the ``best'' candidates from lists of names associated with different racial and gender groups in 40 occupations. Their findings revealed a significant preference for white male names. \citet{an-etal-2024-large} similarly investigated the impact of applicant names on LLM-generated hiring decisions. They manipulated applicant names in the prompts to measure the influence of perceived ethnicity and gender on acceptance rates. Their results showed evidence of discrimination against Hispanic applicants, particularly males. Although these studies effectively demonstrate the presence of name-based biases, they are limited to strictly designed scenarios and may not be generalized to other occupational contexts or broader real-world applications. 

\subsection{Evaluating Bias by LLMs' Word Choices}

Another line of research has focused on analyzing the linguistic features of LLM-generated text to uncover biases~\citep{wan-etal-2023-kelly, 10.1145/3582269.3615599, lee-etal-2024-exploring-inherent, soundararajan-delany-2024-investigating, kumar2024decodingbiasesautomatedmethods, bas2024assessinggenderbiasllms}. For example, \citet{wan-etal-2023-kelly} analyzed gender biases in LLM-generated recommendation letters by examining the language style and lexical content used to describe male and female candidates. They revealed significant gender bias in ChatGPT and Alpaca, with female candidates often described using more communal descriptions (e.g., kind, great to work with) while male candidates were described using more agentic descriptions (e.g., expert, a standout in the industry). \citet{10.1145/3582269.3615599} used ambiguous sentences with gendered occupations and pronouns to assess LLMs' reliance on gender stereotypes and analyze their reasoning. They found that LLMs frequently rely on gender stereotypes to resolve pronouns, revealing a bias in the models' understanding of gender and occupations. \citet{lee-etal-2024-exploring-inherent} investigated the inherent biases and toxicity of LLMs in the Korean social context. They assign various personas to LLMs and evaluate the toxicity of word choices. They found that specific persona-issue combinations could produce harmful content, indicating that LLMs can reflect and amplify social biases existing in cultural contexts.

Unlike previous approaches, we prompt LLMs to generate stories about a variety of occupations and analyze the gender ratio of the characters in the stories. 
This method allows LLMs to generate open-ended stories rather than responding to restricted scenarios (e.g.,~\citep{nghiem-etal-2024-gotta, an-etal-2024-large}), eliminates the need for meticulously designed prompts (e.g.,~\citep{10.1145/3582269.3615599}), and requires no semantic analysis of the output's content (e.g.,~\citep{wan-etal-2023-kelly}).
Thus, it has the advantage of capturing natural and authentic biases embedded in the model's narrative style, providing a less constrained perspective on occupational gender stereotypes.

\section{Evaluating Occupational Stereotypes} \label{sec:eval-method}

This section details four key stages of our methodology: (1) selecting occupations, (2) asking LLMs to generate story openings and settings, (3) classifying the gender of the main character in each story, and (4) analyzing the results for gender bias.

We selected Llama 3.2 3B~\citep{meta2024llama}, Gemma 2 (2B, 9B and 27B)~\citep{gemmateam2024gemma2improvingopen}, GPT-4o, GPT-4o mini~\citep{openai2024gpt4ocard}, Gemini 1.5 Flash and Gemini 1.5 Flash-8B, Gemini 1.5 pro and Gemini 2.0 flash~\citep{geminiteam2024gemini15unlockingmultimodal} for evaluation because these models strike a balance between accessibility and representational capabilities, allowing us to conduct comprehensive evaluations without requiring complex infrastructure. Specifically, we built Llama 3.2 3B and Gemma 2 (2B, 9B, and 27B) via Ollama, while the GPT-4o, GPT-4o mini, Gemini 1.5 Flash, Gemini 1.5 Flash-8B, Gemini 1.5 Pro, and Gemini 2.0 Flash models were accessed through official APIs. All models were evaluated using default settings, as this is the configuration that most users experience.

\subsection{Occupation Selection}

To create a list of occupations for evaluation, we combined two sources: the U.S.~Bureau of Labor Statistics (BLS)~\citep{bls_cpsaat11}, which reflects the current occupational landscape in the United States, and the gender stereotype rating study (GSR)~\citep{Kennison2003}, which represents human perceptions of gender stereotypes associated with occupations. The integration process involved matching occupations across the two datasets. However, BLS data often grouped multiple related occupations into broader categories. For example, BLS defines ``Writers and Authors'' as a single category, while GSR assigns separate ratings for ``Writer'' and ``Author.'' To reconcile these differences, we map occupations from GSR to their closest corresponding categories in BLS, keeping the finer distinctions from GSR wherever possible. We eventually obtain a set of 106 occupations. %

The GSR dataset contains the participants' ratings on perceived gender associations of occupations, measured on a scale of 1 to 7 (1 represents the 0\% male association; 7 represents 100\% male association). We converted these ratings to a male association percentage:
\begin{equation}
\text{Male Association Rate} = \frac{(\text{Rating} - 1)}{6} \times 100\%
\label{eq:male_assoc}
\end{equation}

\subsection{Prompt Design and Story Generation}

We ask LLMs to generate stories by the prompt:

\vspace{0.5cm}
\textit{Please provide a story opening and setting, including the full name of the main character, who is a [occupation].}
\vspace{0.5cm}

For these 106 occupations, we initiated 75 independent conversation sessions with each of the 10 LLMs, prompting them to generate one story per session. This approach ensures that each story is created in isolation, preventing any influence or carryover effects from other stories about the same occupation. 
The generated stories were substantial in length, with a mean of 1388.02 words.

\subsection{Gender Classification of Story Characters}
We chose to infer gender from character names rather than pronouns because a pronoun-based approach was unreliable for two reasons. First, in stories with multiple characters, it is difficult to confirm whether pronouns such as ``he'' or ``she'' refer to the protagonist. Second, some narratives were written in the first person; they lacked third-person pronouns for the main character. 
Eventually, we extracted the character's name from each story and utilized the \texttt{nomquamgender} package~\citep{van2023open} to classify the gender of the names. We set 0.49 to 0.51 as neutral; predictions in the range were excluded from further analysis. We sampled 3000 stories from the LLM-generated corpus and manually labeled the gender of the characters for these stories. We found that it achieved an accuracy of $98.77\%$, which we deemed sufficient for the following analyses.

\subsection{Bias Analysis}

For each occupation in a given LLM, we analyzed the gender distribution of characters in the generated stories, using it as an indicator of the model's tendencies to associate genders with occupations.

To evaluate the gender stereotypes exhibited by each LLM, we compared the gender distribution against two benchmarks: (1) the stereotype gender associations~\citep{Kennison2003}, reflecting human perceptions of gender stereotypes for occupations; (2) the gender distribution in the U.S.~labor force~\citep{bls_cpsaat11}, representing real-world demographics. These comparisons assess whether LLMs' gender biases align with societal perceptions, real-world statistics, or diverge in ways that introduce new biases.

\begin{figure}[tb]
    \centering
    \includegraphics[width=.8\columnwidth]{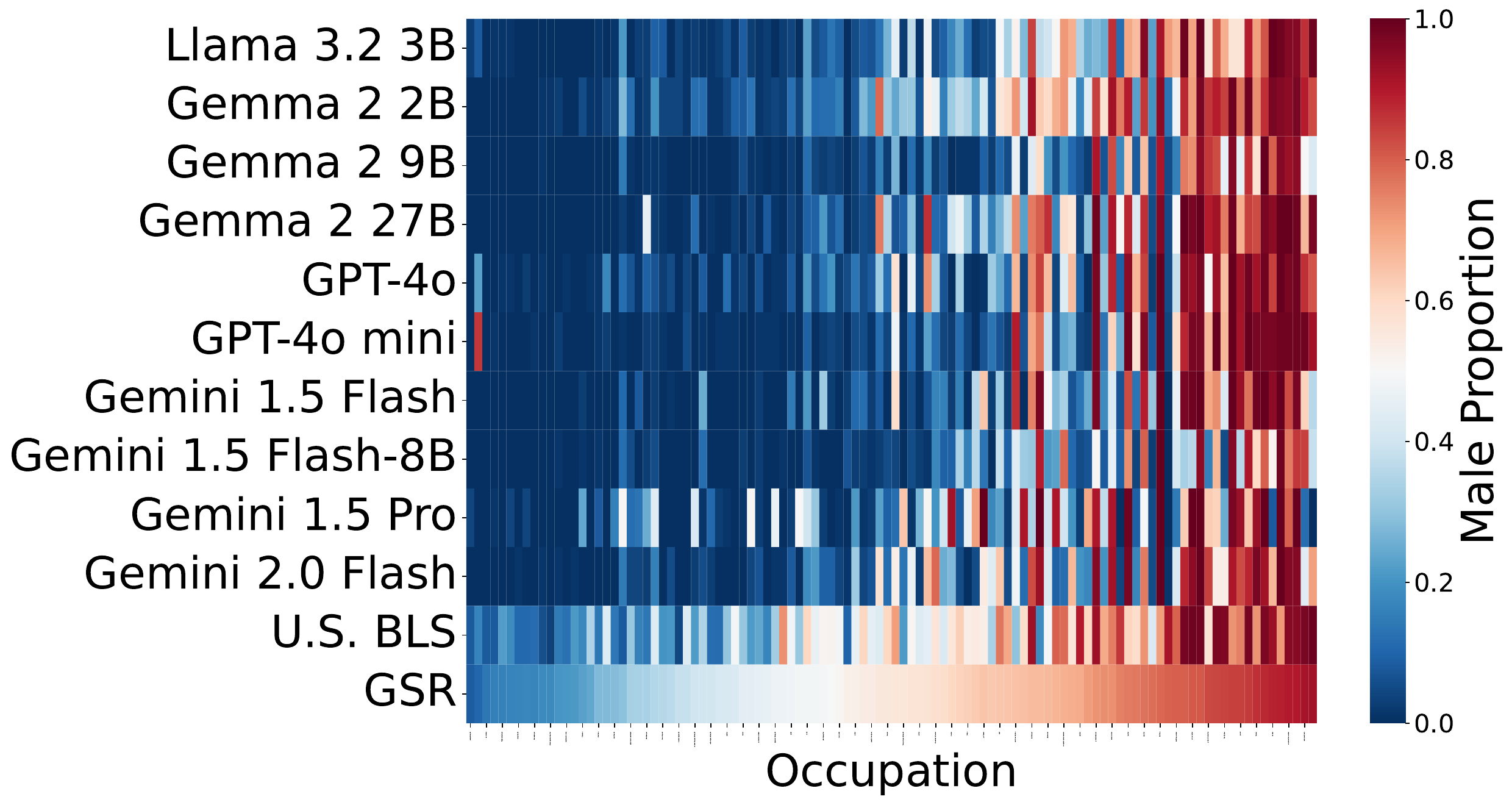}
    \caption{Occupation Gender Heatmap. The occupations are ranked from the most female-oriented (left) to the most male-oriented (right) according to the GSR. Our analysis of 106 occupations revealed LLMs' significant gender skew. Female characters predominated ($\geq$80\% of stories) in 35 occupations, whereas male characters did so in only 5 of 106.}
    \vspace{-15pt}%
    \label{fig:heatmap-order-by-human}
\end{figure}

\begin{figure}[tb]
    \centering
    \includegraphics[width=.62\columnwidth]{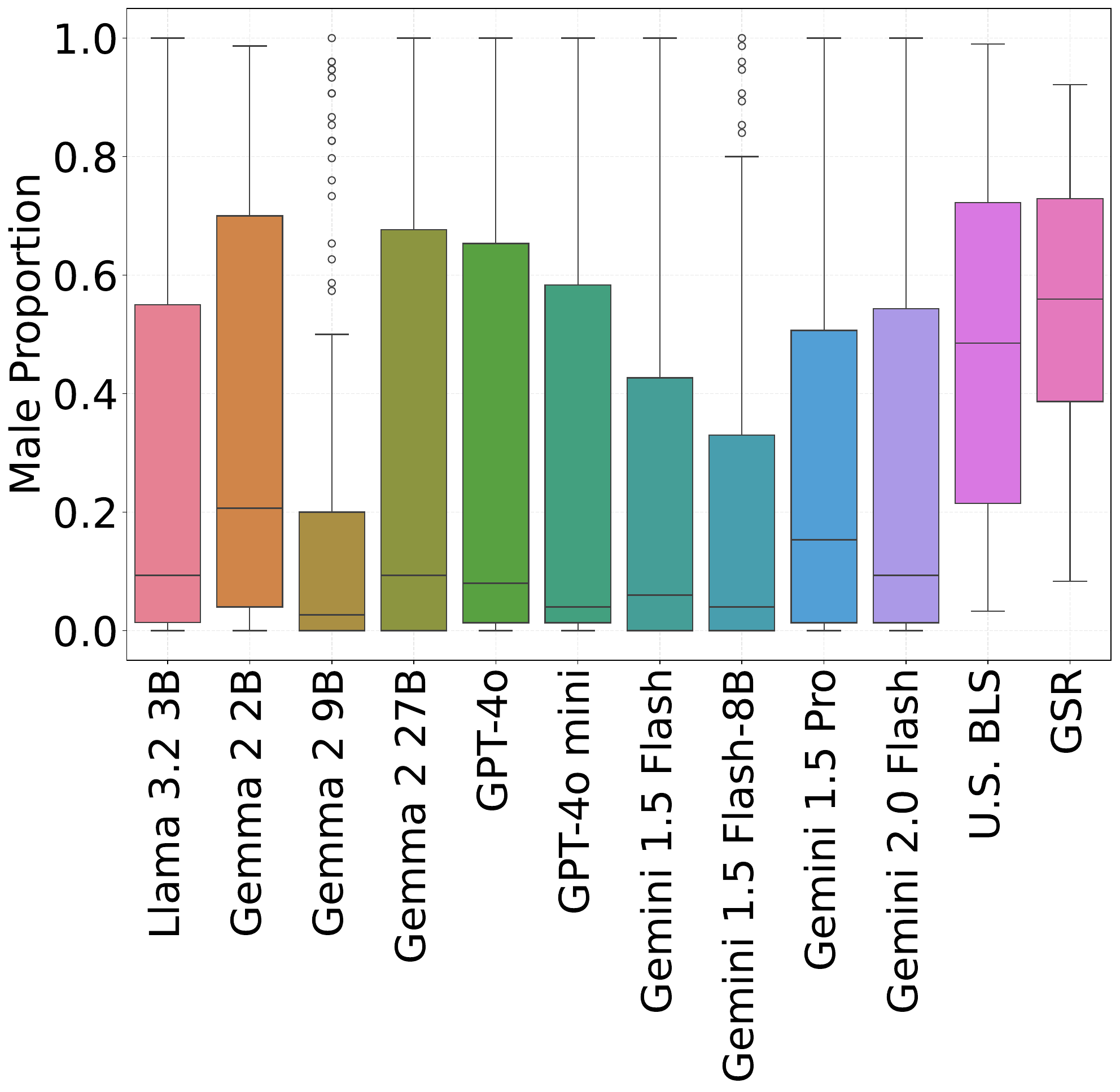}
    \caption{The boxplot of the male ratio for all occupations generated by each LLM and two benchmarks. All LLMs exhibit a significant bias towards female representation, with median male proportions consistently below 20\%.}
    \vspace{-10pt}%
    \label{fig:profession-gender-boxplot}
\end{figure}

\begin{table}[t]
\centering
\caption{Kendall's Tau and Averaged Male Proportion Deviations between LLMs and two benchmarks U.S.~BLS and GSR. A higher value of Kendall's Tau means more consistent; a lower value of Averaged Male Proportion Deviation means a bias towards females. All Kendall's Tau results are significant ($p$-values $\ll 10^{-5}$).}
\begin{tabular}{l|cc|cc}
\toprule
& \multicolumn{2}{c|}{Kendall's Tau} & \multicolumn{2}{c}{Averaged Deviations} \\
\cmidrule(lr){2-3} \cmidrule(lr){4-5}
Model & U.S. BLS & GSR & U.S. BLS & GSR \\
\midrule
Llama3.2 3B & 0.6352 & \textbf{0.6975} & \textbf{-0.1999} & -0.2537 \\
Gemma2 2B & 0.6642 & \textbf{0.7411} & \textbf{-0.1399} & -0.1938 \\
Gemma2 9B & 0.6125 & \textbf{0.6919} & \textbf{-0.2843} & -0.3382 \\
Gemma2 27B & 0.6159 & \textbf{0.7201} & \textbf{-0.1720} & -0.2259 \\
Gpt-4o mini & 0.5939 & \textbf{0.6712} & \textbf{-0.2178} & -0.2717 \\
Gpt-4o & 0.5375 & \textbf{0.5666} & \textbf{-0.1899} & -0.2438 \\
Gemini 1.5 flash & 0.5522 & \textbf{0.6364} & \textbf{-0.2253} & -0.2792 \\
Gemini 1.5 flash-8B & 0.5577 & \textbf{0.6368} & \textbf{-0.2918} & -0.3456 \\
Gemini 1.5 pro & 0.3368 & \textbf{0.4557} & \textbf{-0.1753} & -0.2292 \\
Gemini 2.0 flash & 0.5383 & \textbf{0.6362} & \textbf{-0.1950} & -0.2489 \\
\bottomrule

\end{tabular}
\label{tab:kendall-deviation-to-benchmark} %
\end{table}

\section{Discovery} \label{sec:discovery}

Our experiments revealed notable and complex patterns in LLMs regarding gender representation in occupational narratives.

\subsection{Pervasive Overrepresentation of Females}
\label{subsec:female_overrepresentation}

A primary discovery is the significant overrepresentation of female characters generated by the 10 contemporary LLMs evaluated. Analysis of 7950 story generations per model shows a strong skew. This trend is observed in all modern LLMs tested, with female character generation rates typically ranging from 65\% to more than 80\%. 
This shows a substantial skew compared to the median male proportion in U.S.~BLS data (47.3\% male) and GSR (56.5\% male).

Figure~\ref{fig:heatmap-order-by-human} ranks occupations based on human ratings, from those perceived as the most female-oriented to the most male-oriented based on GSR. The heatmap shows that all LLMs tested show a significant tendency toward female characters. This pronounced trend has direct consequences at the occupational level: of the 106 occupations analyzed, 35 featured female characters in at least $80\%$ of the generated stories, while only 5 occupations exhibited similar dominance for male characters. Figure~\ref{fig:profession-gender-boxplot} further illustrates this gender distribution across all occupations. Although the median gender ratios of real-world data (US BLS) and human evaluations (GSR) are relatively balanced (0.473 and 0.565 male proportion, respectively), all LLMs exhibit a significant bias, with median male proportions falling below $20\%$, and in some cases even below $10\%$.

We quantify the overall bias of each LLM based on the Averaged Male Proportion Deviations (AMPD), calculated as Equation~\ref{eq:avgdev}.
\begin{equation} \label{eq:avgdev}
\text{AMPD}\left(P^{\text{LLM}}, P^{\text{Bench}}\right) = \frac{1}{N_{\text{occ}}} \sum_{i=1}^{N_{\text{occ}}} \left( P_{i}^{\text{LLM}} - P_{i}^{\text{Bench}} \right),
\end{equation}
where $P^{\text{LLM}}$ and $P^{\text{Bench}}$ are vectors containing the male proportions for all $N_{\text{occ}}$ occupations; $P_{i}^{\text{LLM}}$ and $P_{i}^{\text{Bench}}$ are the specific proportions for an individual occupation $i$.

When compared against both U.S. BLS labor statistics and GSR human stereotype ratings, all ten LLMs consistently yield negative AMPD across all 106 occupations, i.e., LLMs generate a lower proportion of male characters than is present in either real-world labor data or human stereotypical perceptions.

Interestingly, LLMs often depict neutral or slightly male-oriented jobs (e.g., emergency medical care, lawyer) as female-dominated. This systemic overrepresentation suggests intentional mitigation strategies or emergent model behaviors leading to a new form of distributional skew. This phenomenon could have significant impacts: while it might challenge existing stereotypes and promote broader gender representation in traditionally male-dominated fields, it also risks creating new stereotypes, oversimplifying social dynamics, or misrepresenting the composition of the workforce.

\subsection{LLM Rankings of Occupational Gender Ratios Align Closer to Human Judgments}
\label{subsec:rankings_vs_human_judgment}

Despite the overall shift towards female character generation, when we rank occupations by their generated male-to-female ratios, a different pattern emerges. In particular, we compare the LLM rankings with the BLS data ranking and the GSR ranking using Kendall's Tau. The result, as indicated in Table~\ref{tab:kendall-deviation-to-benchmark}, shows that the consistency of the ranking between human perceptions (GSR) and LLM-generated rankings is higher than the consistency between BLS data and LLM-generated rankings. This suggests that while the absolute number of female characters is inflated, the relative ordering of occupations from most male-associated to most female-associated still predominantly mirrors societal stereotypes.

\subsection{Female Overrepresentation May Stem from Alignment}
\label{subsec:GPT2-xl}
\begin{figure}[tb]
    \centering
    \includegraphics[width=.6\columnwidth]{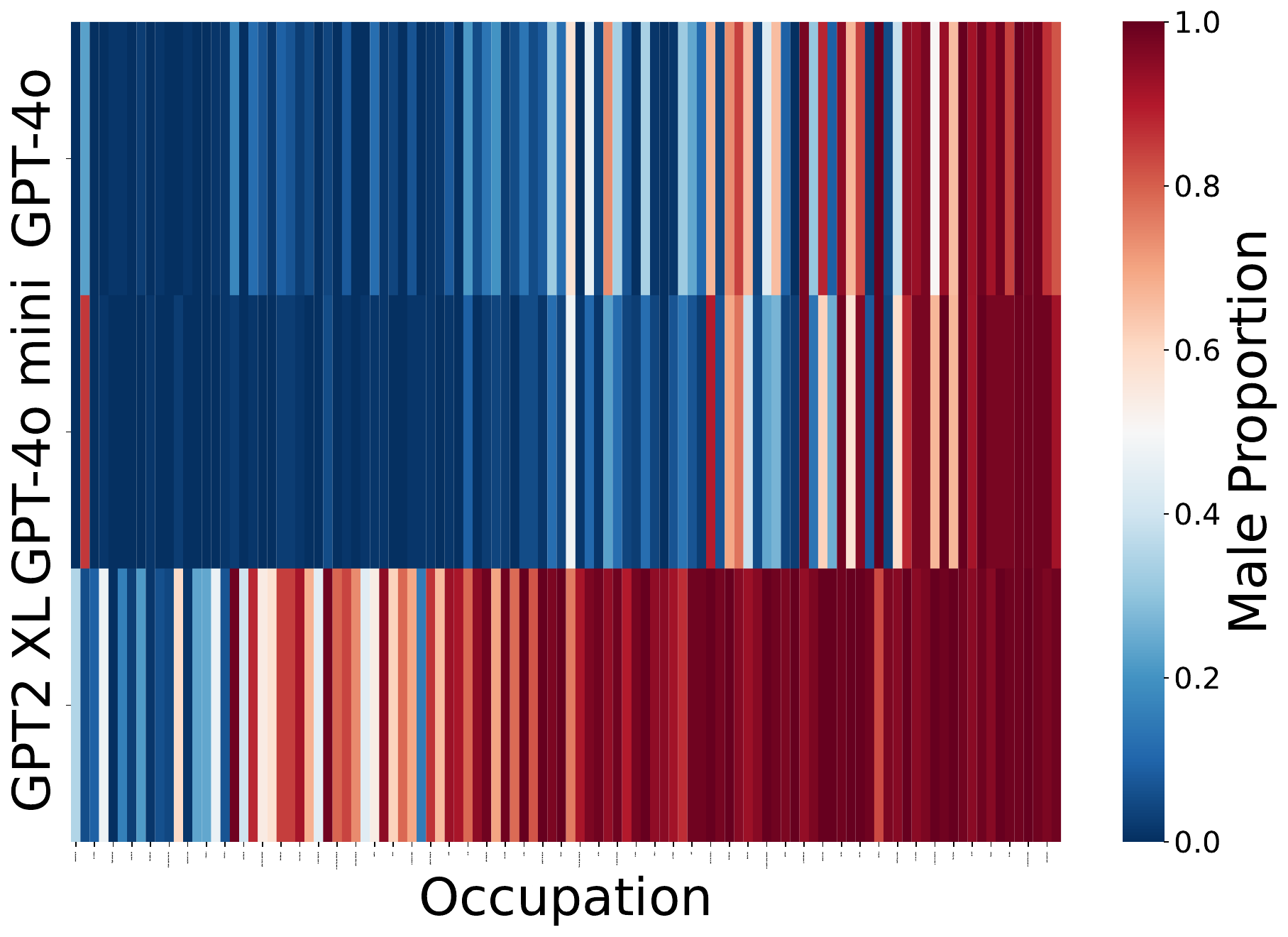}
    \caption{Occupation Gender Heatmap of GPT-4o, GPT-4o-mini and GPT2-XL. With male protagonists appearing in 82.0\% of its stories, GPT2-XL shows a pattern that is completely opposite to that of the SFT- and RLHF-tuned models.}
    \vspace{-15pt}%
    \label{fig:gpt2-xl_profession_gender_heatmap}
\end{figure}

Female characters are frequently overrepresented in contemporary LLMs (as detailed in Section~\ref{subsec:female_overrepresentation}). This has led us to investigate the potential reasons for this imbalance. One possible reason is that this skew is, at least in part, a consequence of alignment strategies, particularly SFT and RLHF, applied during model development. These processes aim to make models more helpful, harmless, and honest, and may include explicit or implicit efforts to address historical underrepresentation or promote balanced gender portrayals.

To investigate the influence of SFT and RLHF, we compare the gender distribution of characters generated by more recent, heavily SFT and RLHF-tuned models (GPT-4o mini and GPT-4o) with that of GPT-2 XL, an LLM developed prior to the widespread adoption of such intensive alignment techniques. Figure~\ref{fig:gpt2-xl_profession_gender_heatmap} provides a visual comparison. As seen, GPT-2 XL exhibits a markedly different pattern -- a more balanced male proportion or even a tendency toward male characters -- unlike the pronounced female skew observed in the SFT- and RLHF-tuned models shown along with it.

This discrepancy implies that strong female overrepresentation may be an emergent property or an intentional outcome after alignment. Developers, in their efforts to mitigate pre-existing biases or achieve diversity goals, might have calibrated the models to over-generate female characters. Although such interventions may aim for increased inclusivity, they risk creating new artificial skews.

\section{Discussion} \label{sec:disc}
Our study sheds light on gender biases in LLMs and their impact on real-world applications. This section explores key implications, possible causes, and fairness considerations in LLM deployment.

First, the overrepresentation of female characters across occupations in LLM-generated stories suggests that developers may attempt to correct historical gender imbalances in training data. Although such efforts are commendable and align with the goals of increasing inclusivity, they may unintentionally lead to overcorrection, which risks fostering new stereotypes, such as the notion that women dominate certain occupations, regardless of real-world gender distributions. These unintended consequences highlight the need for strategies to achieve balance without distorting societal realities or reinforcing counterproductive biases.

Additionally, the disparity between human perception-based rankings and real-world labor statistics underscores the complex interplay between societal norms, training data, and model outputs. LLMs tend to mirror societal perceptions more closely than actual demographics, as evidenced by the alignment of model-generated rankings with human perception-based rankings in general. This raises critical questions about the role of training data and model fine-tuning in shaping the narratives LLMs produce and their potential to either challenge or reinforce existing stereotypes.

Our findings emphasize the ethical responsibilities of LLM developers to ensure fairness and inclusivity while minimizing unintended consequences. These responsibilities extend beyond training processes to include deployment strategies, user education, and ongoing monitoring of model outputs. For instance, users should be informed of the potential biases present in LLM-generated content and encouraged to critically evaluate outputs.

\section*{Acknowledgement and GenAI Usage Disclosure}

We acknowledge support from National Science and Technology Council of Taiwan under grant number 113-2221-E-008-100-MY3. We thank to National Center for High-performance Computing (NCHC) of National Applied Research Laboratories (NARLabs) in Taiwan for providing computational and storage resources.
The authors used ChatGPT and Gemini to improve language and readability. The authors reviewed and edited the content as needed and take full responsibility for the content of the publication.

\bibliographystyle{unsrtnat}
\bibliography{ref}  

@inproceedings{wan-etal-2023-kelly,
    title = "{``}Kelly is a Warm Person, Joseph is a Role Model{''}: Gender Biases in {LLM}-Generated Reference Letters",
    author = "Wan, Yixin  and
      Pu, George  and
      Sun, Jiao  and
      Garimella, Aparna  and
      Chang, Kai-Wei  and
      Peng, Nanyun",
    editor = "Bouamor, Houda  and
      Pino, Juan  and
      Bali, Kalika",
    booktitle = "Findings of the Association for Computational Linguistics: EMNLP 2023",
    month = dec,
    year = "2023",
    address = "Singapore",
    publisher = "Association for Computational Linguistics",
    url = "https://aclanthology.org/2023.findings-emnlp.243",
    doi = "10.18653/v1/2023.findings-emnlp.243",
    pages = "3730--3748",
}

@inproceedings{nghiem-etal-2024-gotta,
    title = "``You Gotta be a Doctor, Lin'' : An Investigation of Name-Based Bias of Large Language Models in Employment Recommendations",
    author = "Nghiem, Huy  and
      Prindle, John  and
      Zhao, Jieyu  and
      Daum\'e Iii, Hal",
    editor = "Al-Onaizan, Yaser  and
      Bansal, Mohit  and
      Chen, Yun-Nung",
    booktitle = "Proceedings of the 2024 Conference on Empirical Methods in Natural Language Processing",
    month = nov,
    year = "2024",
    address = "Miami, Florida, USA",
    publisher = "Association for Computational Linguistics",
    url = "https://aclanthology.org/2024.emnlp-main.413/",
    doi = "10.18653/v1/2024.emnlp-main.413",
    pages = "7268--7287"
}

@inproceedings{an-etal-2024-large,
    title = "Do Large Language Models Discriminate in Hiring Decisions on the Basis of Race, Ethnicity, and Gender?",
    author = "An, Haozhe  and
      Acquaye, Christabel  and
      Wang, Colin  and
      Li, Zongxia  and
      Rudinger, Rachel",
    editor = "Ku, Lun-Wei  and
      Martins, Andre  and
      Srikumar, Vivek",
    booktitle = "Proceedings of the 62nd Annual Meeting of the Association for Computational Linguistics (Volume 2: Short Papers)",
    month = aug,
    year = "2024",
    address = "Bangkok, Thailand",
    publisher = "Association for Computational Linguistics",
    url = "https://aclanthology.org/2024.acl-short.37",
    doi = "10.18653/v1/2024.acl-short.37",
    pages = "386--397",
}

@misc{haim2024whatsnameauditinglarge,
      title={What's in a Name? Auditing Large Language Models for Race and Gender Bias}, 
      author={Amit Haim and Alejandro Salinas and Julian Nyarko},
      year={2024},
      eprint={2402.14875},
      archivePrefix={arXiv},
      primaryClass={cs.CL},
      url={https://arxiv.org/abs/2402.14875}, 
}

@inproceedings{van2023open,
  title={An open-source cultural consensus approach to name-based gender classification},
  author={Van Buskirk, Ian and Clauset, Aaron and Larremore, Daniel B},
  booktitle={Proceedings of the International AAAI Conference on Web and Social Media},
  volume={17},
  pages={866--877},
  year={2023}
}

@inproceedings{10.1145/3582269.3615599,
author = {Kotek, Hadas and Dockum, Rikker and Sun, David},
title = {Gender bias and stereotypes in Large Language Models},
year = "2023",
isbn = {9798400701139},
publisher = {Association for Computing Machinery},
address = {New York, NY, USA},
url = {https://doi.org/10.1145/3582269.3615599},
doi = {10.1145/3582269.3615599},
booktitle = {Proceedings of The ACM Collective Intelligence Conference},
pages = {12-24},
numpages = {13},
location = {Delft, Netherlands},
series = {CI '23}
}

@inproceedings{Salinas_2023,
author = {Salinas, Abel and Shah, Parth and Huang, Yuzhong and McCormack, Robert and Morstatter, Fred},
title = {The Unequal Opportunities of Large Language Models: Examining Demographic Biases in Job Recommendations by ChatGPT and LLaMA},
year = {2023},
isbn = {9798400703812},
publisher = {Association for Computing Machinery},
address = {New York, NY, USA},
url = {https://doi.org/10.1145/3617694.3623257},
doi = {10.1145/3617694.3623257},
booktitle = {Proceedings of the 3rd ACM Conference on Equity and Access in Algorithms, Mechanisms, and Optimization},
articleno = {34},
numpages = {15},
location = {Boston, MA, USA},
series = {EAAMO '23}
}

@article{Kennison2003,
  author    = {Shelia M. Kennison and Jessie L. Trofe},
  title     = {Comprehending Pronouns: A Role for Word-Specific Gender Stereotype Information},
  journal   = {Journal of Psycholinguistic Research},
  year      = {2003},
  volume    = {32},
  number    = {3},
  pages     = {355-378},
  issn      = {1573-6555},
  doi       = {10.1023/A:1023599719948},
  url       = {https://doi.org/10.1023/A:1023599719948},
  id        = {Kennison2003}
}

@inproceedings{lee-etal-2024-exploring-inherent,
    title = "Exploring Inherent Biases in {LLM}s within {K}orean Social Context: A Comparative Analysis of {C}hat{GPT} and {GPT}-4",
    author = "Lee, Seungyoon  and
      Kim, Dong  and
      Jung, Dahyun  and
      Park, Chanjun  and
      Lim, Heuiseok",
    editor = "Cao, Yang (Trista)  and
      Papadimitriou, Isabel  and
      Ovalle, Anaelia  and
      Zampieri, Marcos  and
      Ferraro, Francis  and
      Swayamdipta, Swabha",
    booktitle = "Proceedings of the 2024 Conference of the North American Chapter of the Association for Computational Linguistics: Human Language Technologies (Volume 4: Student Research Workshop)",
    month = jun,
    year = "2024",
    address = "Mexico City, Mexico",
    publisher = "Association for Computational Linguistics",
    url = "https://aclanthology.org/2024.naacl-srw.11/",
    doi = "10.18653/v1/2024.naacl-srw.11",
    pages = "93--104"
}

@misc{bls_cpsaat11,
  author = {{U.S. Bureau of Labor Statistics}},
  title = {Employed persons by detailed occupation, sex, race, and Hispanic or Latino ethnicity},
  url = {https://www.bls.gov/cps/cpsaat11.htm},
  year = "2023", 
}

@inproceedings{soundararajan-delany-2024-investigating,
    title = "Investigating Gender Bias in Large Language Models Through Text Generation",
    author = "Soundararajan, Shweta  and
      Delany, Sarah Jane",
    editor = "Abbas, Mourad  and
      Freihat, Abed Alhakim",
    booktitle = "Proceedings of the 7th International Conference on Natural Language and Speech Processing (ICNLSP 2024)",
    month = oct,
    year = "2024",
    address = "Trento",
    publisher = "Association for Computational Linguistics",
    url = "https://aclanthology.org/2024.icnlsp-1.42/",
    pages = "410--424"
}

@misc{bas2024assessinggenderbiasllms,
      title={Assessing Gender Bias in LLMs: Comparing LLM Outputs with Human Perceptions and Official Statistics}, 
      author={Tetiana Bas},
      year={2024},
      eprint={2411.13738},
      archivePrefix={arXiv},
      primaryClass={cs.CL},
      url={https://arxiv.org/abs/2411.13738}, 
}

@misc{kumar2024decodingbiasesautomatedmethods,
      title={Decoding Biases: Automated Methods and LLM Judges for Gender Bias Detection in Language Models}, 
      author={Shachi H Kumar and Saurav Sahay and Sahisnu Mazumder and Eda Okur and Ramesh Manuvinakurike and Nicole Beckage and Hsuan Su and Hung-yi Lee and Lama Nachman},
      year={2024},
      eprint={2408.03907},
      archivePrefix={arXiv},
      primaryClass={cs.CL},
      url={https://arxiv.org/abs/2408.03907}, 
}

@misc{gemmateam2024gemma2improvingopen,
      title={Gemma 2: Improving Open Language Models at a Practical Size}, 
      author={Team, Gemma and Riviere, Morgane and Pathak, Shreya and Sessa, Pier Giuseppe and Hardin, Cassidy and others},
      year={2024},
      eprint={2408.00118},
      archivePrefix={arXiv},
      primaryClass={cs.CL},
      url={https://arxiv.org/abs/2408.00118}, 
}

@misc{geminiteam2024gemini15unlockingmultimodal,
      title={Gemini 1.5: Unlocking multimodal understanding across millions of tokens of context},
      author={Team, Gemini and Georgiev, Petko and Lei, Ving Ian and Burnell, Ryan and others},
      journal={arXiv preprint arXiv:2403.05530},
      year={2024}
}

@misc{openai2024gpt4ocard,
      title={GPT-4o System Card}, 
      author={OpenAI and Hurst, Aaron and Lerer, Adam and Goucher, Adam P. and others},
      year={2024},
      eprint={2410.21276},
      archivePrefix={arXiv},
      primaryClass={cs.CL},
      url={https://arxiv.org/abs/2410.21276}, 
}

@misc{meta2024llama,
  title        = {Llama 3.2: Revolutionizing edge AI and vision with open, customizable models},
  author       = {{Meta AI}},
  year         = {2024},
  url          = {https://ai.meta.com/blog/llama-3-2-connect-2024-vision-edge-mobile-devices/},
}

\end{document}